\documentclass[10pt,twocolumn,letterpaper]{article}

\usepackage{cvpr}              

\definecolor{cvprblue}{rgb}{0.21,0.49,0.74}
\usepackage[pagebackref,breaklinks,colorlinks,allcolors=cvprblue]{hyperref}
\usepackage{multirow}
\usepackage[most]{tcolorbox}
\tcbuselibrary{breakable}
\usepackage{colortbl}
\usepackage{pifont}
\usepackage{makecell}
\usepackage{cuted}

\usepackage{newfloat}
\usepackage{scrextend}
\usepackage{listings}
\usepackage{CJKutf8}

\def\paperID{2498} 
\def\confName{CVPR}
\def\confYear{2025}

\newcommand{\methodabbr}{MBQ}

\title{\methodabbr: Modality-Balanced Quantization for Large Vision-Language Models}

\author{%
Shiyao Li\thanks{Equal contribution. $^{\dag}$Corresponding to: foxdoraame@gmail.com, jiaxt@buaa.edu.cn, yu-wang@tsinghua.edu.cn}$^{~ 1,2}$,
Yingchun Hu$^{* 2,3}$,
Xuefei Ning$^{\dag1}$,
Xihui Liu$^{5}$, 
Ke Hong$^{1,2}$, 
Xiaotao Jia$^{\dag3,4}$, 
Xiuhong Li$^{2}$,
\\
Yaqi Yan$^{6}$, 
Pei Ran$^{6}$, 
Guohao Dai$^{7,2}$, 
Shengen Yan$^{2}$, 
Huazhong Yang$^{1}$, 
Yu Wang$^{\dag1}$
\\
{\normalsize
$^{1}$Tsinghua University
$^{2}$Infinigence-AI
$^{3}$Beihang University
$^{4}$Qingdao Research Institute, Beihang University
}
\\
{\normalsize
$^{5}$University of Hong Kong
$^{6}$Chinatower
$^{7}$Shanghai Jiao Tong University
}
}

\begin{document}
\begin{CJK}{UTF8}{gbsn}
\maketitle

\newtcolorbox[auto counter,crefname={Ex.}{Exs.}]{convexample}[2][]{enhanced,title={Example~\thetcbcounter. \textbf{#1}},colback=green!5!white, colframe=green!50!black, colbacktitle=green!60!black, coltitle=white, #2}

\lstset{
  language=Python,
  basicstyle=\ttfamily\scriptsize,
  breaklines=true,
  tabsize=4,
  numbers=left,
  moredelim=[is][\color{blue}]{|}{|}
}

\newcommand{\tconvexample}[9]{
\begin{convexample}[#1]{#9}
\textbf{Question:}
\input{#2} 

\begin{center}
\includegraphics[width=\textwidth,height=7cm,keepaspectratio]{#3} 
\end{center}

\tcbline
\textbf{FP16 Response:}
\input{#4} 

\tcbline
\textbf{#5 Response:}
\input{#6} 

\tcbline
\textbf{#7 Response:}
\input{#8} 

\end{convexample}
}

\newcommand{\tconvimgexample}[9]{
\begin{convexample}[#1]{#9}
\textbf{Question:}
\input{#2} 

\begin{center}
\includegraphics[width=\textwidth,height=4cm,keepaspectratio]{#3} 
\end{center}

\tcbline
\textbf{FP16 Response:}
\begin{center}
\includegraphics[width=\textwidth,height=4cm,keepaspectratio]{#4} 
\end{center}

\tcbline
\textbf{#5 Response:}
\input{#6} 

\tcbline
\textbf{#7 Response:}
\begin{center}
\includegraphics[width=\textwidth,height=4cm,keepaspectratio]{#8} 
\end{center}

\end{convexample}
}

\begin{abstract}
Vision-Language Models (VLMs) have enabled a variety of real-world applications.
The large parameter size of VLMs brings large memory and computation overhead which poses significant challenges for deployment.
Post-Training Quantization (PTQ) is an effective technique to reduce the memory and computation overhead.
Existing PTQ methods mainly focus on large language models (LLMs), without considering the differences across other modalities.
In this paper, we discover that there is a significant difference in sensitivity between language and vision tokens in large VLMs.
Therefore, treating tokens from different modalities equally, as in existing PTQ methods, may over-emphasize the insensitive modalities, leading to significant accuracy loss.
To deal with the above issue, we propose a simple yet effective method, \textbf{M}odality-\textbf{B}alanced \textbf{Q}uantization (\textbf{\methodabbr{}}), for large VLMs.
Specifically, \methodabbr{} incorporates the different sensitivities across modalities during the calibration process to minimize the reconstruction loss for better quantization parameters.
Extensive experiments show that \methodabbr{} can significantly improve task accuracy by up to 4.4\% and 11.6\% under W3A16 and W4A8 quantization for 7B to 70B VLMs, compared to SOTA baselines.
Additionally, we implement a W3A16 GPU kernel that fuses the dequantization and GEMV operators, achieving a 1.4$\times$ speedup on LLaVA-onevision-7B on the RTX 4090.
The code is available at \url{https://github.com/thu-nics/MBQ}.
\end{abstract}
\section{Introduction}
\label{sec:intro}
Large Vision-Language Models (VLMs) have made significant progress and enabled various real-world tasks, such as image captioning~\cite{coco-cap}, visual question answering (VQA)~\cite{vqa}, and so on.
However, due to the large memory and computation overhead, existing large VLMs, such as LLaVA~\cite{llava}, InternVL~\cite{internvl}, and QwenVL~\cite{qwen-vl}, are hard to deploy on commonly used accelerators, such as GPUs.
For example, the largest LLaVA-onevision VLM, with 72B parameters, requires 144GB of memory, which exceeds the 80GB memory capacity of the A100 GPU.

\begin{figure}[t]
    \centering
    \includegraphics[width=\linewidth]{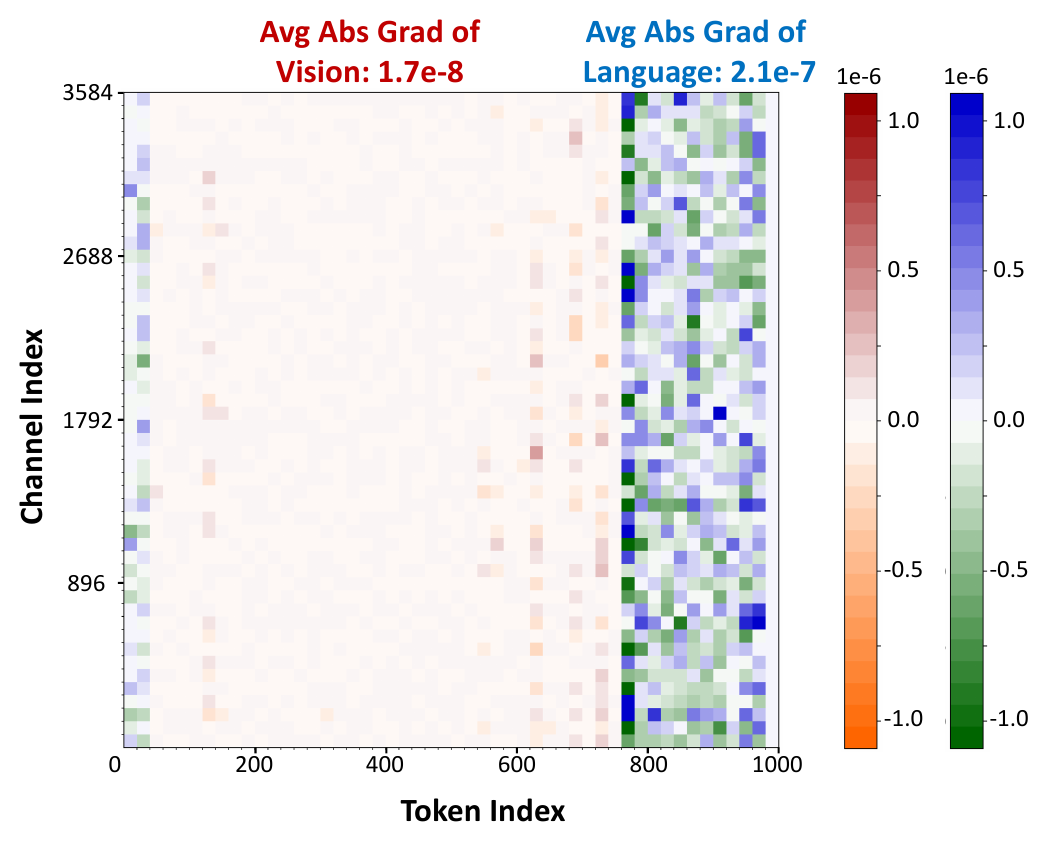}
    \vspace{-20pt}
    \caption{The gradients of loss function w.r.t. the token features of the 13th transformer block in the LLaVA-onevision-7B VLM on COCO caption dataset~\cite{coco-cap,sharegpt4v}. The red and orange represent vision tokens, and the blue and green represent language tokens.}
\label{fig:oracle}
\vspace{-10pt}
\end{figure}

Many methods have already been designed to improve the inference efficiency of LLM`\cite{eff-survey}, including quantization~\cite{awq, gptq,quarot}, sparse attention~\cite{streamingllm,h2o,moa}, efficient decoding strategies~\cite{sot,eagle} and so on.
Among them, Post-Training Quantization (PTQ) methods are effective in reducing the memory access and computation overhead to accelerate LLM inference.
To reduce the memory access and storage overhead, weight-only quantization methods have been developed, such as AWQ~\cite{awq}, GPTQ~\cite{gptq}, QuIP~\cite{quip}, and so on.
To reduce the computation overhead, weight-activation quantization methods, such as SmoothQuant~\cite{smoothquant}, SpinQuant~\cite{spinquant}, FlatQuant~\cite{flatquant}, Atom~\cite{atom}, and so on, have been applied, enabling the use of the faster low-precision tensor cores on GPUs.
To retain the task accuracy, during the calibration process, these methods search for optimal scaling factors~\citep{omniquant, gptq}, channel-wise equalization factors~\citep{awq,smoothquant}, rotation matrices~\citep{spinquant}, etc., by minimizing the feature reconstruction error of each block.

While PTQ methods are well-studied for LLMs, their suitability for VLMs has not been fully explored, where tokens from multiple modalities need to be handled and cross-modality tasks need to be addressed. Our experiments reveal that directly applying SOTA PTQ methods for LLMs to VLMs results in substantial accuracy degradation.
We speculate that the main reason is that existing methods treat vision and language tokens equally, overlooking their significantly different sensitivities.

To verify this, as depicted in \cref{fig:oracle}, using an image-caption pair from the COCO dataset as the input~\cite{coco-cap}, we visualize the loss gradient w.r.t. the output feature of the 13th layer in LLaVA-onevision-7B.
We can see that the average absolute gradient value of the language token features is over 10$\times$ larger than that of vision token features. This 1st-order approximation analysis suggests that with the same size of perturbation on the features, language tokens might impact the loss function more than 10$\times$ as much as vision tokens.
Consequently, treating vision and language tokens equally during calibration may over-emphasize the insensitive vision tokens, resulting in a notable accuracy loss.

Taking the sensitivity differences into consideration, we propose an extremely simple but effective method, \textbf{M}odality-\textbf{B}alanced \textbf{Q}uantization (\textbf{\methodabbr{}}), for quantizing large VLMs.
Specifically, \methodabbr{} uses the gradients of the loss function~\cite{llava} w.r.t. vision and language token features as the sensitivity indicators. 
These sensitivity indicators are then incorporated into the reconstruction loss as the objective for searching optimal channel-wise equalization factors~\cite{awq, smoothquant} in both weight-only (W3A16 and W4A16) and weight-activation quantization (W4A8 and W8A8).
By balancing the effect of different modalities, \methodabbr{} can significantly increase the accuracy of the quantized VLMs.

We conduct extensive experiments across 7B-70B VLMs on challenging vision-language benchmarks.
The results show \methodabbr{} significantly improves the task accuracy by up to 4\% and 11\% under W3A16 and W4A8 quantization compared with other SOTA methods. We also analyze the shortcomings of baseline methods and perform comprehensive ablation studies on various factors, including the choice of calibration datasets, alternative formulations of modality-balancing losses, and the quantization of visual encoders.

For W3A16 quantization, we design a GPU kernel that fuses 3-bit dequantization with general matrix-vector multiplication (GEMV). 
For W4A16, W4A8, and W8A8 quantization, we adopt existing open-source GPU kernels~\cite{qserve, awq} to accelerate the inference process.
Experiments on different workloads show that we can achieve up to 1.4$\times$ end-to-end speedup on LLaVA-onevision-7B on RTX 4090.




\section{Preliminaries}
\label{sec:pre}

\subsection{Quantization Formats}
\label{sec:pre_quant}
In this paper, we focus on applying the uniform integer quantization, which is a commonly used quantization format, to the weight ($\textbf{W}$) and input activation ($\textbf{X}$) matrices of each linear layers in VLMs. 

For weight-only quantization, existing methods typically apply asymmetric uniform quantization for weight groups (i.e., group-wise quantization), as shown below:
\begin{equation}
\label{eq:integer_quant_main}
    \mathbf{W}_{\mathrm{asym}} = \left[ \frac{\mathbf{W}_{\mathrm{FP16}}-Z}{S_{\mathrm{asym}}} \right],
\end{equation}
\begin{equation}
\label{eq:integer_quant_scale}
    S_{\mathrm{asym}} = \frac{\mathrm{max}(\mathbf{W}_{\mathrm{FP16}})-Z}{2^{N}-1},
\end{equation}
where $\mathbf{W}_{\mathrm{FP16}}$ denotes the 16-bit floating-point (FP16) value, $\mathbf{W}_{\mathrm{asym}}$ denotes the low-precision integer value. 
$N$ is bit-width.
$S_{\mathrm{asym}}$ and $Z=\mathrm{min}(\mathbf{W}_{\mathrm{FP16}})$ denote the scaling factor and zero-point of the asymmetric quantization. 

For weight-activation quantization, symmetric uniform quantization is commonly used for both weight and input activation matrices of linear layers:
\begin{equation}
\label{eq:integer_quant_main}
    \mathbf{W}_{\mathrm{sym}} = \left[ \frac{\mathbf{W}_{\mathrm{FP16}}}{S_{\mathrm{sym}}} \right],
\end{equation}
\begin{equation}
\label{eq:integer_quant_scale}
    S_{\mathrm{sym}} = \frac{\mathrm{absmax}(\mathbf{W}_{\mathrm{FP16}})}{2^{N-1}-1},
\end{equation}
where $S_{\mathrm{sym}}$ denotes the scaling factor of the symmetric quantization.

For simplicity, we use $\textbf{WxAy}$ to indicate the quantization format, where the \textbf{x} and \textbf{y} represent to the bit-width for Weight and Activation. 
For example, W4A8 denotes quantizing weights to 4-bit and activations to 8-bit.

\subsection{Channel-Wise Equalization}
\label{sec:pre-cwe}
Existing PTQ methods automatically search for optimal hyperparameters of quantization by minimizing the reconstruction error of each transformer block during a calibration process. 
A series of popular methods~\cite{awq, smoothquant} aim to equalize outliers in weight and activation matrices by channel-wise equalization (CWE).
Specifically, they search the CWE factors \textbf{E} by minimizing the Mean Square Error (MSE) loss in each transformer block.
Taking weight-only quantization as an example, the objective of CWE is shown in the following equation:
\begin{equation}
\label{eq:cwe}
    \mathbf{E^*}=\mathop{\mathrm{arg min}}\limits_{\mathbf{E}}||Q(\mathbf{W}*\mathbf{E})(\mathbf{E}^{-1}*\mathbf{X})-\mathbf{W}\mathbf{X}||^2,
\end{equation}
where $Q$ means the quantization function.

\section{Method}
\label{sec:method}

\subsection{Sensitivity Varies Across Modalities}
\label{sec:method_sen}
As introduced in \cref{sec:pre-cwe}, when using visual-language datasets for calibration, \textbf{CWE treats visual and language activations equally during the calibration process}.

Intuitively, we speculate that the significant performance degradation when applying SOTA LLM quantization methods to VLMs stems from treating different modalities equally. This is because errors in vision tokens might have a smaller impact on the output context compared to introducing the same errors in language tokens, due to the following two reasons:
(1) From the data perspective, visual data generally contains a high degree of redundancy and might be more fault tolerant for small perturbations. 
(2) From the model perspective, \citet{debiasing} discover that the generated content of current VLMs is primarily biased by the pre-trained LLMs rather than the input image.


To verify the above intuition, we evaluate the sensitivity of the output tokens w.r.t the input vision and language tokens on the COCO caption dataset~\cite{coco-cap}.
Specifically, we take the image-caption pairs as inputs of VLMs and calculate gradients of the SFT (Supervised Fine-Tuning) loss function w.r.t. language and vision tokens.
The gradients can reflect the impacts on the output language tokens (caption) when small perturbations are applied to language (caption) and vision (image) token features.
Note that due to the attention mechanism, the gradient of the SFT loss can still backpropagate to vision tokens in each transformer block, even though we only account for the loss of the output language tokens.

As shown in \cref{fig:oracle}, the average absolute gradient of language tokens is an order of magnitude larger than that of vision tokens.
This means that, for a similar perturbation, a vision token's impact on the SFT loss is only 0.1$\times$ that of a language token.
\textbf{Therefore, if we still treat language and vision tokens equally, we will miss the opportunity to leverage the VLM's lower sensitivity to vision tokens to achieve higher performance.}

To demonstrate the impact of accounting for modality-specific sensitivities during calibration, we conduct an oracle experiment by applying a modality-balancing factor of 0.1 to the vision tokens' reconstruction loss during CWE calibration.
The optimization objective, referred to as balanced CWE, is shown below:
\begin{align}
    \mathbf{E^*}=\mathop{\mathrm{arg min}}\limits_{\mathbf{E}}[&||Q(\mathbf{W}*\mathbf{E})(\mathbf{E}^{-1}*\mathbf{X}_l)-\mathbf{W}\mathbf{X}_l||^2 \\
    + 0.1&||Q(\mathbf{W}*\mathbf{E})(\mathbf{E}^{-1}*\mathbf{X}_v)-\mathbf{W}\mathbf{X}_v||^2],
\end{align}
where the $\mathbf{X}_l$ and $\mathbf{X}_v$ mean the language and vision tokens.

\begin{table}[t]
    \centering
    \resizebox{\linewidth}{!}{
    \begin{tabular}{cccccc}
    \toprule
    \multirow{2}{*}{Model} & \multirow{2}{*}{FP16} & \multicolumn{3}{c}{W3} \\
    \cmidrule(lr){3-5}
    ~ & ~ & RTN & CWE & Balanced CWE \\
    \midrule
    LLaVA-ov-7B & 46.00 & 34.67 & 36.56 & \textbf{40.22} \\
    InternVL2-8B & 48.00 & 43.67 & 44.78 & \textbf{46.33} \\
    \bottomrule
    \end{tabular}}
    \caption{The accuracy results (\%) on MMMU benchmark after W3 quantization. The calibration dataset is selected from COCO caption dataset~\cite{coco-cap}. The LLaVA-ov-7B is short for LLaVA-onevision-7B.}
    \label{tab:oracle}
    \vspace{-10pt}
\end{table}

The results of our ablation study are shown in \cref{tab:oracle}, with a heuristic selected modality-balancing factor, the balanced CWE easily surpasses the performance of CWE by 1.55\% $\sim$ 3.66\% under W3 quantization.
The significant improvements indicate the importance of balancing the different sensitivities across different modalities.

\subsection{Modality-Balanced Quantization (\methodabbr{})}
\label{sec:method-mbq}
Given that the sensitivity differences between vision and language tokens may vary across layers and VLM families, exploring an automatic modality-balancing approach could further enhance the performance of the quantized model.

In this section, \textbf{we aim to derive an approach for allocating the optimal Modality-Balanced factors to each layer by directly minimizing the change in the SFT loss function}.
Specifically, we employ the first-order Taylor approximation in \cref{eq:taylor} to determine how the SFT loss $L$ changes in response to a small perturbation $\Delta$ in output activation $\mathbf{Y}$ of each linear layer:
\begin{align}
\label{eq:taylor}
    L(\mathbf{Y}+\Delta) \simeq & L(\mathbf{Y}) + \mathbf{g}^T*\Delta,
\end{align}
where $\mathbf{g}^T$ represents the gradient of the output activation $\mathbf{Y}$.

The change in SFT loss caused by quantization can be expressed as the following equation:
\begin{align}
    ||L(\hat{\mathbf{Y}})|| \simeq & ||\mathbf{g}^T*\Delta||\\
    = & ||\mathbf{g}_v^T*\Delta_v + \mathbf{g}_l^T*\Delta_l||\\
    \leq & ||\mathbf{g}_v^T*\Delta_v|| + ||\mathbf{g}_l^T*\Delta_l||\\
    \leq & \Bar{|\mathbf{g}_v|} * ||\Delta_v|| + \Bar{|\mathbf{g}_l|} * ||\Delta_l||\\
    = & \Bar{|\mathbf{g}_v|} * ||\mathbf{Y}_v-\hat{\mathbf{Y}}_v|| + \Bar{|\mathbf{g}_l|} * ||\mathbf{Y}_l-\hat{\mathbf{Y}}_l||,
\end{align}

where $\mathbf{Y}_v$ and $\mathbf{\hat{Y}}_v$ represent the output vision tokens of the FP16 linear layer and the quantized linear, respectively. 
$\mathbf{Y}_l$ and $\mathbf{\hat{Y}}_l$ represent the output language tokens of the FP16 linear layer and the quantized linear, respectively. 
The $\Bar{|\mathbf{g}_v|}$ and $\Bar{|\mathbf{g}_l|}$ represent the average absolute gradient of each linear layer's output vision and language tokens, respectively.

\textbf{In this paper, we combine \methodabbr{} with channel-wise equalization to search optimal equalization factors for better performance.}

Firstly, in order to accelerate the prefill stage of VLMs, we quantize both the weights and input activations of each linear layer to leverage fast low-precision tensor cores. 
The objective is shown in the following:
\begin{align}
    \mathop{\mathrm{min}}\limits_{\mathbf{E}}[&\Bar{|\mathbf{g}_v|}*||\mathbf{W}\mathbf{X}_v-Q(\mathbf{W}*\mathbf{E})Q(\mathbf{E}^{-1}*\mathbf{X}_v)|| \\
    + &\Bar{|\mathbf{g}_l|}*||\mathbf{W}\mathbf{X}_l-Q(\mathbf{W}*\mathbf{E})Q(\mathbf{E}^{-1}*\mathbf{X}_l)||],
\end{align}
where $\mathbf{X}_v$ and $\mathbf{X}_l$ represent the input token and language activations of each linear layer, respectively.

Secondly, to accelerate the decode stage of VLMs, we only quantize weights to reduce the memory overhead, aiming to minimize the balanced reconstruction error by the following objective:
\begin{align}
    \mathop{\mathrm{min}}\limits_{\mathbf{E}}[&\Bar{|\mathbf{g}_v|}*||\mathbf{W}\mathbf{X}_v-Q(\mathbf{W}*\mathbf{E})(\mathbf{E}^{-1}*\mathbf{X}_v)|| \\
    + &\Bar{|\mathbf{g}_l|}*||\mathbf{W}\mathbf{X}_l-Q(\mathbf{W}*\mathbf{E})(\mathbf{E}^{-1}*\mathbf{X}_l)||],
\end{align}

Notably, unlike directly using the heuristically selected MSE-based balanced CWE loss in \cref{sec:method_sen}, our derived reconstruction error loss function relies on Mean Absolute Error (MAE).
Our ablation study in \cref{sec:exp_ablation} demonstrates that minimizing MAE-based reconstruction loss in \methodabbr{} yields better results than using an MSE-based one.

\subsection{End-to-End Acceleration Implementation}
\label{sec:imple}
As illustrated in \cref{sec:pre_inference}, we need to quantize not only the VLMs with large parameter counts but also the ViT encoders for efficient deployment, which have substantial computational overhead when transforming high-resolution images into vision tokens.
The ViT encoders take a series of image patches as input and generate a set of vision tokens as output.
The input activation of each linear layer in ViT encoders are 2D matrices, so the main computation operator is GEMM.

To this end, we apply weight-activation quantization to the ViT encoders and weight-only quantization for VLMs to accelerate the auto-regressive decode stage.
To achieve practical hardware acceleration for VLMs, we designed a custom fused W3 quantization GPU kernel by fusing the dequantization process with the GEMV operator.
Specifically, we pack eight 3-bit weights into three bytes, and the fused W3 kernel first loads the W3 weights instead of FP16 weights to reduce memory access overhead.
It then dequantizes the W3 weights to FP16.
Finally, the fused W3 kernel performs computations using the FP16 tensor core.

With the fused W3 kernel and open-source GPU kernel libraries~\cite{qserve, awq} for quantization, we can accelerate the inference speed of both the quantized ViT encoders and VLMs.
Detailed experiments in \cref{sec:hardware} demonstrate the hardware performance of the proposed W3 kernel and end-to-end speedups across various scenarios.
\begin{table*}[htbp]
  \centering
  \resizebox{0.8\textwidth}{!}{
      \begin{tabular}{cccccccccc}
        \toprule
        Model & Bitwidth & Method & MMMU & SEED & OCRBench & VizWiz & ScienceQA & TextVQA & Average ($\uparrow$) \\
        \midrule
        \multirow{8}{*}{LLaVA-onevision-7B} & FP16 & - & 46.0 & 74.9 & 62.2 & 60.4 & 85.4 & 76.1 & 67.5\\
        \cmidrule(lr){2-10}
        ~ & \multirow{4}{*}{W3A16} & RTN & 34.7 & 5.9 & 35.9 & 59.2 & 86.2 & 60.9 & 47.1\\
        ~ & ~ & GPTQ & 41.9 & \textbf{72.9} & 55.7 & 56.4 & \textbf{86.4} & 71.3 & 64.1\\
        ~ & ~ & AWQ & 36.6 & 53.0 & 59.3 & 58.5 & 83.2 & 73.0 & 60.6\\
        ~ & ~ & \methodabbr{} & \textbf{42.0} & 69.7 & \textbf{61.1} & \textbf{60.7} & 85.0 & \textbf{73.3} & \textbf{65.3}\\
        \cmidrule(lr){2-10}
        ~ & \multirow{3}{*}{W4A8} & RTN & 38.2 & 50.3 & 40.1 & 58.2 & 88.3 & 61.5 & 56.1\\
        ~ & ~ & SQ & 30.9 & 42.7 & 32.0 & 56.7 & 87.1 & 56.9 & 51.1\\
        ~ & ~ & \methodabbr{} & \textbf{42.6} & \textbf{67.7} & \textbf{52.3} & \textbf{58.9} & \textbf{88.5} & \textbf{68.3} & \textbf{63.1}\\
    
        \midrule
    
        \multirow{8}{*}{InternVL2-8B} & FP16 & - & 48.0 & 76.0 & 76.5 & 61.1 & 96.2 & 77.0 & 72.5\\
        \cmidrule(lr){2-10}
        ~ & \multirow{4}{*}{W3A16} & RTN & 43.7 & 74.9 & 74.0 & 56.0 & 95.6 & 74.6 & 69.8\\
        ~ & ~ & GPTQ & 41.7 & 73.4 & 70.2 & \textbf{59.9} & 89.5 & 73.1 & 68.0\\
        ~ & ~ & AWQ & 44.8 & 75.2 & 74.7 & 58.9 & 95.5 & 74.2 & 70.6\\
        ~ & ~ & \methodabbr{} & \textbf{46.9} & \textbf{75.4} & \textbf{75.1} & 58.7 & \textbf{95.6} & \textbf{75.1} & \textbf{71.1}\\
        \cmidrule(lr){2-10}
        ~ & \multirow{3}{*}{W4A8} & RTN & 44.3 & 74.0 & 72.0 & \textbf{57.1} & 95.5 & \textbf{73.1} & 69.3\\
        ~ & ~ & SQ & 41.8 & 73.8 & 70.9 & 54.2 & 95.1 & 72.6 & 68.1\\
        ~ & ~ & \methodabbr{} & \textbf{45.6} & \textbf{74.3} & \textbf{73.0} & 56.5 & \textbf{95.8} & 72.3 & \textbf{69.6}\\
    
        \midrule
        
        \multirow{8}{*}{Qwen2-VL-7B} & FP16 & - & 50.6 & 76.4 & 80.7 & 68.3 & 85.1 & 82.0 & 73.8\\
        \cmidrule(lr){2-10}
        ~ & \multirow{4}{*}{W3A16} & RTN & 44.9 & 74.8 & 60.0 & 65.2 & 81.5 & 71.2 & 66.3\\
        ~ & ~ & GPTQ & 43.1 & 73.7 & 74.8 & 64.3 & 79.7 & 76.7 & 68.7\\
        ~ & ~ & AWQ & 44.7 & \textbf{75.1} & \textbf{76.9} & \textbf{68.0} & 82.5 & 79.5 & 71.1\\
        ~ & ~ & \methodabbr{} & \textbf{47.9} & 74.8 & 76.8 & 67.7 & \textbf{82.8} & \textbf{79.9} & \textbf{71.6}\\
        \cmidrule(lr){2-10}
        ~ & \multirow{3}{*}{W4A8} & RTN & 43.8 & 74.9 & 60.3 & 58.9 & 78.9 & 71.0 & 64.6\\
        ~ & ~ & SQ & 45.9 & 75.0 & 57.1 & 52.3 & 80.9 & 68.2 & 63.2\\
        ~ & ~ & \methodabbr{} & \textbf{47.2} & \textbf{75.1} & \textbf{72.8} & \textbf{59.3} & \textbf{81.2} & \textbf{75.0} & \textbf{68.4}\\
        \bottomrule
      \end{tabular}
    }
  \caption{Main results on the small models of LLaVA-onevision, InternVL2, and Qwen2-VL families. ``SQ'' is short for SmoothQuant.}
  \label{tab:main-small}
\end{table*}

\begin{table*}[htbp]
  \centering
  \resizebox{0.8\textwidth}{!}{
      \begin{tabular}{cccccccccc}
        \toprule
        Model & Bitwidth & Method & MMMU & SEED & OCRBench & VizWiz & ScienceQA & TextVQA & Average ($\uparrow$) \\
        \midrule
        \multirow{8}{*}{LLaVA-onevision-72B} & FP16 & - & 56.1 & 78.1 & 73.2 & 69.2 & 90.0 & 79.3 & 74.3 \\
        \cmidrule(lr){2-10}
        ~ & \multirow{4}{*}{W3A16} & RTN & 53.9 & 77.4 & 68.2 & 66.1 & 89.5 & 77.4 & 72.1\\
        ~ & ~ & GPTQ & 52.7 & 76.0 & 69.7 & 68.3 & 89.3 & 77.9 & 72.3 \\
        ~ & ~ & AWQ & 33.4 & 71.2 & 48.7 & 49.3 & 69.2 & 58.8 & 55.1\\
        ~ & ~ & \methodabbr{} & \textbf{54.4} & \textbf{77.6} & \textbf{71.6} & \textbf{69.0} & \textbf{90.3} & \textbf{78.5} & \textbf{73.6}\\
        \cmidrule(lr){2-10}
        ~ & \multirow{3}{*}{W4A8} & RTN & 54.8 & \textbf{76.6} & \textbf{64.5} & 64.7 & 89.0 & \textbf{74.5} & 70.7\\
        ~ & ~ & SQ & 51.6 & 76.6 & 64.2 & 65.7 & 89.1 & 74.4 & 70.3\\
        ~ & ~ & \methodabbr{} & \textbf{55.6} & 76.5 & 64.4 & \textbf{65.7} & \textbf{89.2} & 73.3 & \textbf{70.8}\\
    
        \midrule
    
        \multirow{8}{*}{InternVL2-26B} & FP16 & - & 47.1 & 76.8 & 77.9 & 66.2 & 97.5 & 82.1 & 74.6\\
        \cmidrule(lr){2-10}
        ~ & \multirow{4}{*}{W3A16} & RTN & 46.6 & 75.7 & 75.9 & \textbf{64.7} & 96.4 & 80.6 & 73.3\\
        ~ & ~ & GPTQ & 44.8 & 75.8 & 76.0 & 60.9 & 96.3 & 80.1 & 72.3 \\
        ~ & ~ & AWQ & 46.4 & 76.2 & 76.4 & 64.5 & 96.7 & 81.0 & 73.5\\
        ~ & ~ & \methodabbr{} & \textbf{47.1} & \textbf{76.3} & \textbf{76.5} & 64.5 & \textbf{97.3} & \textbf{81.1} & \textbf{73.8}\\
        \cmidrule(lr){2-10}
        ~ & \multirow{3}{*}{W4A8} & RTN & \textbf{44.7} & \textbf{76.0} & 76.4 & \textbf{62.6} & 96.7 & 79.6 & 72.7\\
        ~ & ~ & SQ & 38.2 & 70.6 & 68.5 & 56.7 & 86.3 & 72.6 & 65.5\\
        ~ & ~ & \methodabbr{} & 44.0 & 75.7 & \textbf{77.5} & 62.0 & \textbf{97.1} & \textbf{80.0} & \textbf{72.7}\\

        \midrule
        
        \multirow{8}{*}{Qwen2-VL-72B} & FP16 & - & 61.1 & 77.6 & 79.9 & 76.0 & 91.6 & 82.5 & 78.1\\
        \cmidrule(lr){2-10}
        ~ & \multirow{4}{*}{W3A16} & RTN & 57.7 & 77.5 & 70.4 & 74.8 & 89.7 & 79.7 & 75.0\\
        ~ & ~ & GPTQ & 57.3 & 77.2 & 78.5 & 73.6 & \textbf{91.5} & 81.6 & 76.6 \\
        ~ & ~ & AWQ & 59.6 & 77.6 & \textbf{79.6} & 75.4 & 90.4 & 82.4 & 77.5\\
        ~ & ~ & \methodabbr & \textbf{59.6} & \textbf{77.7} & 79.4 & \textbf{75.6} & 90.5 & \textbf{82.5} & \textbf{77.6}\\
        \cmidrule(lr){2-10}
        ~ & \multirow{3}{*}{W4A8} & RTN & \textbf{58.1} & \textbf{76.6} & 66.2 & 71.3 & \textbf{90.1} & 77.0 & 73.2\\
        ~ & ~ & SQ & 55.9 & 76.4 & 65.5 & 69.7 & 88.8 & 76.9 & 72.2\\
        ~ & ~ & \methodabbr & 57.7 & 76.3 & \textbf{77.5} & \textbf{73.6} & 89.6 & \textbf{80.5} & \textbf{75.8}\\
        
        \bottomrule
      \end{tabular}
    }
  \caption{Main results on the large models of LLaVA-onevision, InternVL2, and Qwen2-VL families. ``SQ'' is short for SmoothQuant.}
  \label{tab:main-large}
\end{table*}

\begin{table*}[t]
    \centering
    \resizebox{0.9\textwidth}{!}{
        \begin{tabular}{cccccccc}
            \toprule
            \multirow{3}{*}{BitWidth} & \multirow{3}{*}{Method} & \multicolumn{4}{c}{Components} & \multirow{3}{*}{MMMU $(\uparrow)$} & \multirow{3}{*}{SEED $(\uparrow)$} \\
            \cmidrule(lr){3-6}
            ~ & ~ & \makecell{Channel-wise\\Equalization} & \makecell{COCO\\Calib.} & \makecell{Modality-Balance\\(MSE)} & \makecell{Modality-Balance\\(MAE)} & ~ & ~ \\

            \midrule
            
            FP16 & - & - & - & - & - & 46.0 & 71.1 \\
            
            \midrule
            
            \multirow{5}{*}{W3A16} & RTN & \ding{55} & \ding{55} & \ding{55} & \ding{55} & 34.7 & 10.4 \\
            ~ & AWQ & \ding{51} & \ding{55} & \ding{55} & \ding{55} & 36.6 & 51.5\\
            ~ & - & \ding{51} & \ding{51} & \ding{55} & \ding{55} & 38.7 & 61.8\\
            ~ & - & \ding{51} & \ding{51} & \ding{51} & \ding{55} & 40.8 & 64.8\\
            \cmidrule(lr){2-8}
            ~ & \methodabbr{} & \ding{51} & \ding{51} & \ding{55} & \ding{51} & 42.0 & 66.4 \\
            
            \midrule
            
            \multirow{5}{*}{W4A8} & RTN & \ding{55} & \ding{55} & \ding{55} & \ding{55} & 38.2 & 48.4 \\
            ~ & SQ & \ding{51} & \ding{55} & \ding{55} & \ding{55} & 30.9 & 41.6 \\
            ~ & - & \ding{51} & \ding{51} & \ding{55} & \ding{55} & 29.2 & 10.2\\            
            ~ & - & \ding{51} & \ding{51} & \ding{51} & \ding{55} & 41.9 & 63.5\\
            \cmidrule(lr){2-8}
            ~ & \methodabbr{} & \ding{51} & \ding{51} & \ding{55} & \ding{51} & 42.6 & 64.4 \\
            \bottomrule
        \end{tabular}
    }
    \caption{The ablation study on LLaVA-onevision-7B with W3A16 and W4A8 quantization. ``SQ'' is short for SmoothQuant.}
    \label{tab:ablation}
    \vspace{-10pt}
\end{table*}

\section{Experiments}
\label{sec:exp}

\subsection{Experimental Setups}
\label{sec:exp_setups}

\subsubsection{Calibration Datasets}
The calibration task requires leveraging both the VLM's ability to understand image details and its language modeling capabilities.
``Image captioning'' is one of the tasks that addresses both aspects.
Specifically, we choose the improved COCO Caption dataset proposed by ShareGPT4V~\cite{sharegpt4v} and sample 128 image-caption pairs as the calibration dataset.
\citet{sharegpt4v} use GPT4-V to generate a high-quality caption for each image.
For each VLM, we apply the corresponding conversation template to each image-caption pair to create an instructional format.

\subsubsection{Evaluation Datasets}
To evaluate the performance of the quantized model, we conduct experiments on various vision-language benchmarks based on the LMMs-Eval~\cite{lmmseval}.
Specifically, we use OCRBench~\cite{ocrbench} and TextVQA~\cite{textvqa} for text recognition and comprehension, VizWiz~\cite{vizwiz} and SEED-Bench~\cite{seed-bench} to test visual perception, and ScienceQA~\cite{scienceqa} and MMMU~\cite{mmmu} to evaluate visual reasoning.
Additionally, to demonstrate the practical conversational performance of the quantized VLM, we present several cases on the LLaVA-bench-in-the-wild~\cite{llava} and LLaVA-bench-wilder~\cite{llavanext} datasets in the supplementary \cref{sec:supp-case}.

\subsubsection{Models}
We benchmark various quantization methods on LLaVA-onevision~\cite{llava-onevision}, InternVL2~\cite{internvl}, and Qwen2-VL~\cite{qwen-vl} families. 
For each model family, we select both a smaller and a larger parameter version to showcase the capability of \methodabbr{} across different model sizes.
For LLaVA-onevision, we select models with 7B and 72B parameters, which utilize Qwen2-7B/-72B for the VLM and SigLIP-400M~\cite{siglip} for the ViT encoder.
For InternVL2, we evaluate the 8B and 26B models, which incorporate InternLM2-8B/-20B as the VLM component and use InternViT-300M/-6B as the vision encoder.
For Qwen2-VL, we use models with 7B and 72B parameters in our evaluation, which use Qwen2-7B/-72B for the VLM component and a 675M ViT encoder.

\subsubsection{Baselines}
For weight-only quantization, we compare \methodabbr{} with the vanilla round-to-nearest quantization (RTN) and AWQ~\cite{awq} under W3A16, which is based on channel-wise equalization.
We apply group-wise asymmetric quantization for each method and keep the group size at 128.
For weight-activation quantization, we compare with both the RTN and SmoothQuant~\cite{smoothquant} under W4A8, which also relies on channel-wise equalization.
As discussed in SmoothQuant, we apply per-token symmetric quantization for activations and per-channel (output) symmetric quantization for weights to take advantage of low-precision tensor cores.
The evaluation results of W4 and W8A8 are shown in Supplementary \cref{sec:supp-ae}.

\subsection{Main Results}
\label{sec:exp_main}


\textbf{For Weight-only Quantization, as shown in \cref{tab:main-small} and \cref{tab:main-large}, the results of RTN quantization indicate that smaller VLMs are more sensitive to weight-only quantization.}
The average accuracy loss for RTN-quantized 7B and 8B VLMs is 9.6\%, significantly higher than the 1.5\% loss seen in VLMs over 26B.
This trend also aligns with observations in LLMs~\cite{qllm-eval,zeroquantv2}.

The proposed \methodabbr{} can significantly outperform the AWQ baseline across different families.
Especially within the LLaVA-onevision family, \methodabbr{} achieves an average accuracy improvement of 4.2\% on the 7B VLM and 18.5\% on the 72B VLM compared to AWQ. 
It is worth noting that for the LLaVA-onevision-72B VLM, AWQ even shows a 17\% performance drop compared to RTN. 
Instead, the proposed \methodabbr{} can significantly improve the average accuracy and surpass RTN quantization.
For the InternVL2 and Qwen2-VL families, \methodabbr{} can also outperform the RTN and AWQ baselines by around 1\%.

\textbf{For Weight-Activation Quantization, similar to weight-only quantization, \methodabbr{} can significantly outperform SmoothQuant and RTN baselines}, with improvements of up to 11.6\%.
In many cases, the average performance of SmoothQuant is lower than that of RTN quantization, especially in InternVL2-26B under W4A8.
The results indicate that with activation quantization, our insight into modality-balancing becomes more crucial for improving the performance of quantized VLMs, as the core idea of modality-balancing mainly addresses the sensitivity differences among various modalities within the activations.

\subsection{Ablation Study}
\label{sec:exp_ablation}

\subsubsection{The Effect of Calibration Datasets}
\label{sec:abla_calib_effect}
As AWQ and SmoothQuant are designed for LLM quantization, they use the Pile~\cite{thepile} validation dataset during calibration, which only contains language data.

Directly apply the vision-language dataset as calibration does not consistently improve the performance of the quantized VLMs.
As shown in \cref{tab:ablation}, when we use the COCO caption dataset as the calibration dataset, the performance of AWQ W3A16 can significantly increase by 2.1\% and 10.3\% on MMMU and SEED datasets, respectively.
However, the performance of SmoothQuant W4A8 with the COCO caption calibration dataset significantly \textbf{decreases} by 1.7\% and 31.4\% on MMMU and SEED datasets.

We speculate that this is because weight-activation quantization requires considering both weight and activation quantization errors.
Ignoring the sensitivity differences between vision and language tokens in activations leads to a significant performance drop in SmoothQuant, even falling behind RTN quantization.

\subsubsection{The Effect of Modality-Balance}
\label{sec:abla_mb_effect}
Modality-Balancing plays a crucial role in weight-activation quantization and can also improve the performance of weight-only quantization.
As shown in \cref{tab:ablation}, with the help of Modality-Balancing (MAE), the SmoothQuant with COCO calibration can significantly improve 13.4\% and 54.2\% on MMMU and SEED, respectively.
For the weight-only quantization, Modality-Balancing (MAE) can increase the accuracy by 3.3\% and 4.6\% on MMMU and SEED, respectively.
This significant performance improvement confirms the importance of accounting for the varying sensitivities of different modalities during the calibration process.

We find that Modality-Balancing (MAE) can consistently outperform Modality-Balancing (MSE) in both weight-only and weight-activation quantization. 
For both W3A16 and W4A8, Modality-Balancing with MAE reconstruction loss can achieve over 1\% accuracy improvement on both MMMU and SEED datasets.
Therefore, we recommend using the MAE reconstruction loss form, derived directly from the gradient of the activation with respect to the SFT loss, rather than the MSE reconstruction loss.

In addition, we also study two different reweight strategies to show the effectiveness of MBQ:
(1) We randomly divided calibration tokens into two equal groups to reweight quant error. For W3A16 LLaVA-onevision-7B, the accuracy on OCRBench is 1.3\% lower than MBQ, which is similar to the case without reweighting.
(2) We directly use the gradient of each token to perform token-wise quantization error reweighting.
For W3A16 LLaVA-onevision-7B, the accuracy on OCRBench is 1.5\% lower than MBQ.
We find that this is because \textbf{the gradient of many vision tokens is zero}, leading to an inability to account for the quantization error of these tokens.
We will highlight these new results.

\subsubsection{Combine \methodabbr{} with Rotation-based Quantization}
\label{sec:abla_rot_quant}
From \cref{tab:main-large}, we observe that for large VLMs, the performance improvement of \methodabbr{} over RTN weight-activation quantization is not substantial, and there remains a significant performance gap compared to FP16 VLMs.

As demonstrated in \cref{sec:method}, the proposed Modality-Balancing approach can be seamlessly integrated with any block-wise LLM quantization method to support VLMs.
To this end, we combine Modality-Balancing with the SOTA rotation-based quantization technique~\cite{flatquant}, resulting in \methodabbr{} (Rot), which can achieve significant better performance.
As illustrated in \cref{tab:rotation}, across three different large VLMs, the proposed \methodabbr{} (Rot) under W4A8 quantization exhibits an accuracy loss of less than 1.1\% compared to the original FP16 VLMs.
Furthermore, when compared to other state-of-the-art methods, W4A8 \methodabbr{} (Rot) achieves performance improvements of up to 9.2\%.

\begin{table}[t]
  \centering
  \resizebox{\linewidth}{!}{
      \begin{tabular}{cccccc}
        \toprule
        Model & Bitwidth & Method & MMMU & OCRBench & VizWiz \\
        \midrule
        ~ & FP16 & - & 56.1 & 73.2 & 69.2 \\
        \cmidrule(lr){2-6}
        LLaVA-one & ~ & QuaRot & 53.0 & 70.8 & 67.9 \\
        vision-72B & W4A8 & SpinQuant & 52.6 & 70.9 & 61.0 \\
        ~ & ~ & \methodabbr{} (Rot) & \textbf{55.8} & \textbf{72.5} & \textbf{69.0} \\
    
        \midrule
    
        ~ & FP16 & - & 47.1 & 77.9 & 66.2 \\
        \cmidrule(lr){2-6}
        InternVL2 & ~ & QuaRot & 44.2 & 76.7 & 62.1 \\
        -26B & W4A8 & SpinQuant & 43.7 & 76.8 & 64.5 \\
        ~ & ~ & \methodabbr{} (Rot) & \textbf{47.4} & \textbf{77.8} & \textbf{65.6} \\
    
        \midrule
        
        ~ & FP16 & - & 61.1 & 79.9 & 76.0 \\
        \cmidrule(lr){2-6}
        Qwen-VL & ~ & QuaRot & 49.7 & 78.1 & 71.4 \\
        -72B & W4A8 & SpinQuant & 50.8 & 77.4 & 69.3 \\
        ~ & ~ & \methodabbr{} (Rot) & \textbf{60.0} & \textbf{79.9} & \textbf{75.4} \\
        \bottomrule
      \end{tabular}
    }
  \caption{Rotation-based quantization results on the large models of LLaVA-onevision, InternVL2, and Qwen2-VL families.}
  \label{tab:rotation}
  \vspace{-10pt}
\end{table}

\subsubsection{Quantize Both Visual Encoder and VLM}
\label{sec:abla_sys_prof}
In \cref{sec:imple}, we analyze the different efficiency bottlenecks of the ViT encoder and VLM.
In real-world applications, we need to quantize both components for higher acceleration ratios.
As shown in \cref{tab:vit_quant}, we quantize the ViT encoder with SmoothQuant and VLM with \methodabbr{} in LLaVA-onevision-7B and evaluate the accuracy on MMMU and VizWiz benchmarks.

Experimental results show that applying W4A8 quantization to the ViT encoder does not lead to a significant performance drop; instead, it even improves performance on some benchmarks.
This suggests that the ViT encoder is not particularly sensitive to quantization, possibly due to the redundancy in vision tokens discussed in \cref{sec:intro}. 
Therefore, quantizing the ViT encoder alongside the VLM quantization is feasible from an algorithmic performance perspective.

\begin{table}
  \centering
  \resizebox{0.8\linewidth}{!}{
      \begin{tabular}{cccc}
        \toprule
            ViT & VLM & MMMU $(\uparrow)$ & VizWiz $(\uparrow)$ \\ 
            \midrule
            FP16 & FP16 & 46.0 & 60.4 \\
            \midrule
            FP16 & W3A16 & 42.0 & 60.7 \\ 
            FP16 & W4A8 & 42.6 & 58.9 \\ 
            \midrule
            W4A8 & W3A16 & 42.6 & 62.8 \\ 
            W4A8 & W4A8 & 43.1 & 60.1 \\ 
        \bottomrule
      \end{tabular}
    }
  \caption{The results of the quantized ViT encoder and VLM in LLaVA-onevision-7B on the MMMU and VizWiz. We use \methodabbr{} for VLM quantization and SmoothQuant for ViT quantization.}
  \label{tab:vit_quant}
\end{table}

\subsubsection{The Performance on Language-only Benchmark}
The main idea of the proposed \methodabbr{} is to consider the sensitivity across different modalities during quantization, aiming to enhance performance in both vision-language and language-only tasks.
Accordingly, we evaluated the performance of the quantized LLaVA-onevision-7B VLM on the MMLU benchmark with different quantization methods.

As shown in \cref{tab:language-bench}, with W3A16 and W4A8 quantization, MBQ achieves a performance improvement of 0.9\% and 2\%, compared to AWQ and SmoothQuant.
This result demonstrates that for quantized VLMs, considering the sensitivity differences across modalities not only significantly reduces the performance loss on vision-language tasks, but also helps maintain performance on language-only tasks.

\begin{table}
    \centering
    \resizebox{0.75\linewidth}{!}{
    \begin{tabular}{ccc}
        \toprule
        BitWidth & Method & MMLU $(\uparrow)$ \\
        \midrule
        FP16 & - & 65.9 \\

        \midrule        
        \multirow{3}{*}{W3A16} & AWQ (Pile) & 61.9 \\
        ~ & AWQ (COCO) & 62.0 \\
        ~ & \methodabbr{} & \textbf{62.9} \\
        
        \midrule
        \multirow{3}{*}{W4A8} & SQ (Pile) & 59.8 \\
        ~ & SQ (COCO) & 59.0 \\
        ~ & \methodabbr{} & \textbf{61.8} \\
        
        \bottomrule
    \end{tabular}}
    \caption{The results of quantized LLaVA-onevision-7B on the MMLU benchmark. ``SQ'' is short for SmoothQuant.}
    \label{tab:language-bench}
    \vspace{-10pt}
\end{table}

\subsection{Efficiency Evaluation}
\label{sec:hardware}

\subsubsection{Single Kernel Performance}
We evaluate the speed of the proposed fused W3A16 GEMV kernel on the RTX 4090 GPU, testing linear layers with different weight matrix shapes in LLaVA-onevision-7B.
Specifically, as shown in \cref{tab:single_kernel}, we evaluate the kernel on 4 different shapes of weight matrices, including 3584 $\times$ 3584 (the out\_proj layers), 3584$\times$10752 (the qkv\_proj layers), 3584$\times$18944 (the up\_proj and gate\_proj layers), and 18944$\times$3584 (the down\_proj layers).
For each shape, we compare the proposed fused W3A16 GEMV kernel with the original FP16 GEMV kernel.

The evaluation results show that the fused W3A16 GEMV kernel achieves a speedup of 1.2$\times$ to 5.0$\times$ across 4 different shapes, compared to the FP16 GEMV baseline.
As the matrix size increases, the fused W3A16 kernel achieves a more significant speedup.
This is because GEMV performance is memory-bound, and the impact of memory access speed becomes more pronounced with larger weight matrices.
By applying W3A16 quantization, memory access latency is significantly reduced in large matrix GEMV operations, leading to a greater speedup.

\begin{table}[]
    \centering
    \resizebox{\linewidth}{!}{
    \begin{tabular}{cccc}
        \toprule
        Shape (IC $\times$ OC) & FP16 (ms) & W3A16 (ms) & Speedup\\
        \midrule
        3584 $\times$ 3584 & 3.0 & 2.6 & 1.2$\times$ \\
        3584 $\times$ 10752 & 8.3 & 2.7 & 3.1$\times$ \\
        3584 $\times$ 18944 & 14.4 & 2.9 & 5.0$\times$ \\
        18944 $\times$ 3584 & 14.6 & 3.1 & 4.7$\times$ \\
        \bottomrule
    \end{tabular}
    }
    \caption{The speed up of the fused W3A16 General Matrix-Vector Multiplication (GEMV) kernel on RTX 4090 GPU. ``IC'' and ``OC'' are short of ``Input Channel'' and ``Output Channel''.}
    \label{tab:single_kernel}
    \vspace{-10pt}
\end{table}


\subsubsection{End-to-End Performance}
With the help of the proposed W3A16 GPU kernel and the W4A8 GPU kernel from Qserve~\cite{qserve}, we evaluate the latency of both the ViT encoder and the VLM in LLaVA-onevision-7B on RTX 4090.
We run both the ViT encoder and VLM with FlashAttention-2~\cite{flashattn2}.
The evaluation results are shown in \cref{tab:e2e_speed}.

\textbf{For the ViT encoder}, the embedding layer will convert each input image into 729 (27 $\times$ 27) tokens as the input.
As discussed in \cref{sec:imple}, the ViT encoder only has the prefill stage, the main operators are GEMMs, which is compute-bound.
In this case, we apply the W4A8 kernel to the ViT encoder and achieve 1.15$\times$ speedup.

\textbf{For the VLM}, as discussed in \cref{sec:pre_inference}, the VLM has two distinct stages, and each stage requires different quantization schemes for inference acceleration. 

Specifically, in order to accelerate the prefill stage, we need to apply weight-activation quantization to use the low-precision tensor cores.
With W4A8 quantization, we evaluate the inference latency of the quantized VLM with the open-source GPU kernel.
As shown in \cref{tab:e2e_speed}, when the input token length is 512 and 1024 tokens, W4A8 can achieve 1.16$\times$ and 1.18$\times$, respectively.
As the token length increases, the speedup gradually becomes greater.

In order to accelerate the decode stage, we apply W3A16 quantization to the VLM.
Since the latency of the decode stage remains similar across different input token lengths, we measure the average latency of the decode stage at input token lengths of 128, 256, 512, and 1024.
As shown in \cref{tab:e2e_speed}, both W4A8 and W3A16 quantization can accelerate the decode stage.
However, the W3A16-quantized VLM is 1.23× faster than the W4A8 model. 
This result primarily stems from two factors:
(1) the W3A16-quantized model has lower memory access overhead for large weight matrices compared to W4A8, and 
(2) the W4A8 operator requires additional dynamic quantization of activations, which incurs additional computational time.

For practical acceleration, we use weight-activation quantization to accelerate the ViT encoder and weight-only quantization to speed up the decode stage of VLMs.
The main reason is that the FP16 VLM's prefill stage takes as long to process 1,024 tokens as the decode stage takes to generate just 4 tokens, as shown in \cref{tab:e2e_speed}.
In real-world applications, the number of tokens to be decoded is large, necessitating multiple forward passes during the decode stage, which is the most time-consuming part of generation tasks.
Thus, weight-only quantization is more suitable for VLMs.

\begin{table}[]
    \centering
    \resizebox{\linewidth}{!}{
    \begin{tabular}{ccccc}
        \toprule
        Model & Stage & FP16 (ms) & W3A16 (ms) & W4A8 (ms) \\
        \midrule
        ViT & Prefill (729 tokens) & 11.2 & - & 9.7 \\
        \midrule
        ~ & Prefill (512 tokens) & 68.8 & - & 59.4 \\
        VLM & Prefill (1024 tokens) & 109.0 & - & 92.7 \\
        ~ & Decode & 29.6 & 21.1 & 26.3 \\
        \bottomrule
    \end{tabular}
    }
    \caption{The end-to-end speed up of LLaVA-onevision-7B on RTX4090 with fused GPU kernels.}
    \label{tab:e2e_speed}
    \vspace{-10pt}
\end{table}
\section{Conclusion}
\label{sec:con}
In this paper, we identify a key phenomenon: the sensitivities of vision and language tokens to quantization differ significantly.
Based on this finding, we propose Modality-Balanced Quantization (\methodabbr{}), a simple but effective quantization method to improve the accuracy of quantized VLMs for both weight-only and weight-activation quantization.
Specifically, we use the gradients of the SFT loss function w.r.t. vision and language tokens as sensitivity indicators to balance the reconstruction loss during calibration.
\methodabbr{} can outperform existing SOTA PTQ methods by 4.4\% and 11.6\% on both W4A8 and W3A16 across various VLMs.
With the proposed W3A16 CUDA kernel, we achieve 1.4$\times$ decoding speedup compared with FP16 baseline.

\noindent\textbf{Acknowledgements.}
This work was supported by the National Natural Science Foundation of China (No. 62325405, 62104128, U19B2019, U21B2031, 61832007, 62204164, 92364201), Tsinghua EE Xilinx AI Research Fund, and Beijing National Research Center for Information Science and Technology (BNRist).
We thank for all the support from Infinigence-AI.

{
    \small
    \bibliographystyle{ieeenat_fullname}
    \bibliography{main}
}


\clearpage
\setcounter{page}{1}
\maketitlesupplementary

\section{Additional Preliminaries}
\begin{table*}[t]
    \centering
    \resizebox{\textwidth}{!}{
    \begin{tabular}{ccccccccc}
        \toprule
        \multirow{2}{*}{Model} & \multirow{2}{*}{Bitwidth} & \multirow{2}{*}{Method} & \multicolumn{5}{c}{Error Types} & Total \\
        \cmidrule(lr){4-8}
        ~ & ~ & ~ & No Output & Randomness & Repetition & Condition Missing & Semantic Error & Bad Cases \\
        
        \midrule
        \multirow{2}{*}{LLaVA-onevision-7B} & \multirow{2}{*}{W4A8} & SQ & 61 & 16 & 15 & 32 & 25 & 149 \\
        ~ & ~ & \methodabbr{} & 3 & 0 & 2 & 3 & 37 & 45 \\

        \midrule
        \multirow{2}{*}{LLaVA-onevision-72B} & \multirow{2}{*}{W3} & AWQ & 30 & 0 & 3 & 11 & 42 & 86 \\
        ~ & ~ & \methodabbr{} & 0 & 0 & 1 & 0 & 37 & 38 \\
        
        \bottomrule
    \end{tabular}
    }
    \caption{The number of samples for the five error types in the LLaVA-bench-in-the-wild~\cite{llava} and LLaVA-bench-wilder~\cite{llavanext} dataset. The total number of samples is \textbf{188}. ``SQ'' is short for SmoothQuant.}
    \label{tab:summary-case}
\end{table*}

\subsection{The Inference Process of VLMs}
\label{sec:pre_inference}
\begin{figure}[t]
    \centering
    \includegraphics[width=\linewidth]{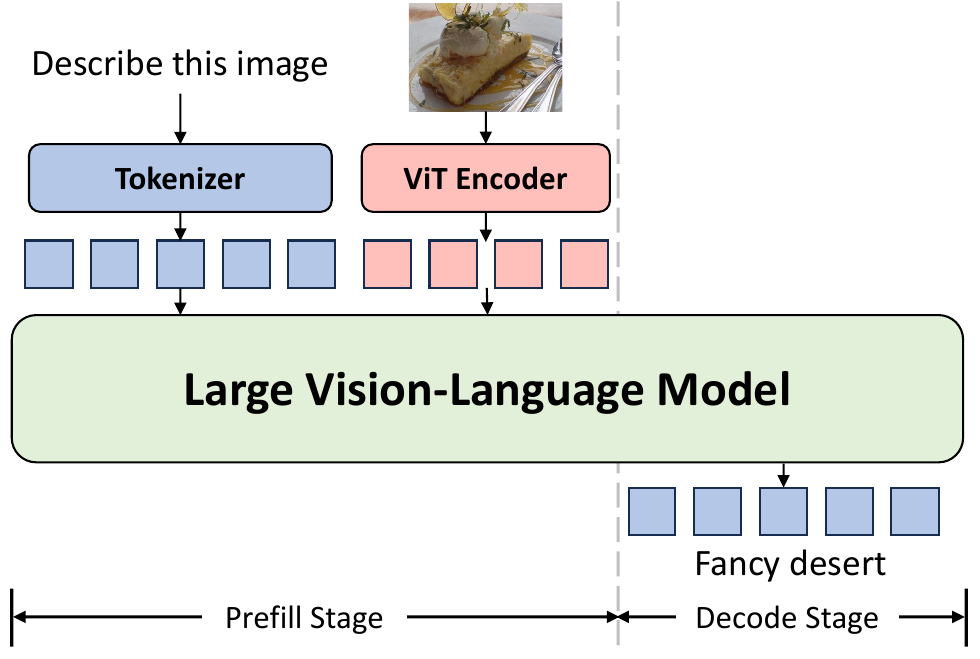}
    \vspace{-0.2cm}
    \caption{The inference process of Large VLMs. The blue patches represent language tokens, the red patches represent vision tokens.}
\label{fig:vlm-inference}
\end{figure}

The inference process of VLMs is shown in \cref{fig:vlm-inference}.
The whole inference system consists of three key components:
\begin{itemize}
    \item \textbf{Language Tokenizer}: Transform natural language sentences into a series of language tokens.
    \item \textbf{ViT Encoder}: Transform images into a series of vision tokens.
    \item \textbf{Large VLM}: Take the language and vision tokens as input, and generate \textbf{language} tokens one by one.
\end{itemize}

Specifically, the transformer-based~\cite{transformer} VLMs have two distinctive stages, including the prefill and the decode stages.
Take batch size = 1 as an example.

\textbf{During the prefill stage}, VLMs can take both vision tokens and language tokens as the input prompt.
In this stage, VLMs aim to understand both the vision and language information and the connections across each modality.
The Key and Value tensors of each attention block in VLMs are stored as the KV Cache to save the computation overhead in the decode stage.
The input activation of each layer is typically a large 2D matrix, \textit{making the primary computation operation in the prefill stage the General Matrix Multiply (GEMM), which is compute-bound}.

\textbf{During the decode stage}, VLMs take one generated token from step $t$ as the input and use the KV Cache to generate the next token of step $t+1$.
The generation of the current token depends on one previously generated token and the KV Cache.
In this case, the input activation of each layer is typically a large 1D vector, and \textit{the main computation operator in the decode stage is the General Matrix-Vector Multiply (GEMV), which is memory-bound}.

\section{Related Work}
\label{sec:rw}

\subsection{LLM Quantization}
\label{sec:rw_quantization}
Post-Training Quantization (PTQ) techniques are widely used in LLMs to accelerate the inference process.
They employ the low-precision data format and computation to reduce the memory and computation overhead.

To accelerate the memory-bound decode stage of LLMs, existing methods apply weight-only quantization to reduce the memory access overhead.
GPTQ~\cite{gptq} quantizes one weight channel at each step and iteratively adjusts the unquantized weights to mitigate reconstruction errors of each transformer block.
AWQ~\cite{awq} searches for proper channel-wise equalization factors by minimizing the block-wise reconstruction loss.
SpQR and LLM-MQ~\cite{spqr, llm-mq} propose mixed-precision quantization to allocate higher precision for weight outliers, while the rest of the weights are quantized to low-precision.
QuIP~\cite{quip} introduces LDLQ, an optimal adaptive method for a quadratic proxy objective.
It reveals that ensuring incoherence between weight and Hessian matrices can enhance the effectiveness of LDLQ. QuIP utilizes LDLQ and achieves incoherence by employing random orthogonal matrix multiplication. 

To accelerate the compute-bound prefill stage of LLMs, existing methods propose to use the weight-activation quantization to leverage faster low-precision tensor cores.
SmoothQuant~\cite{smoothquant} employs a channel-wise equalization technique to address the challenges of quantizing activation values. This method expands the data range of weight channels while shrinking the data range of corresponding activation channels to achieve better data equalization. 
Omniquant~\cite{omniquant} optimizes the boundaries for weight clipping and the scaling factor for equivalent transformation to minimize reconstruction errors block-by-block.
Atom~\cite{atom} employs a strategy involving mixed-precision and dynamic quantization for activations. 
Recently, many studies~\cite{spinquant, quarot} follow the computational invariance idea, by multiplying rotation matrices to the weight matrices and activation matrices.

However, these methods focus solely on a single language modality without considering the differences between tokens from different modalities in multimodal scenarios, which is the core distinction between \methodabbr{} and existing quantization approaches.
It is also worth noting that many existing studies search for various hyperparameters by minimizing reconstruction error, where \methodabbr{} can be used to achieve performance improvements with these methods on VLMs.

\subsection{Efficient VLM}
\label{sec:eff_mllm}
To improve the efficiency of Large Vision-Language Models, existing work primarily focuses on designing lightweight modules, compressing vision tokens, and employing efficient model architectures.

Firstly, for the lightweight model design, an effective approach is to incorporate efficient components within the VLMs.
Some research~\cite{paligemma, llavaphi} directly utilizes pre-trained small language models with fewer than 3B parameters as their backbone, while others~\cite{mobilevlm} train a small language model from scratch. For modality alignment,~\cite{blip2, flamingo} utilizes a lightweight transformer, while subsequent work~\cite{llava, vila, internvl, llava-onevision} directly adopts Linear or MLP layers to align the visual modality with the language model's latent space.

Secondly, the number of vision tokens increases with image resolution, imposing a substantial computational burden on VLMs. To address this issue,~\cite{llava-uhd, llava-prumerge, madtp} propose vision token reduction techniques to significantly lower the number of vision tokens, while some approaches~\cite{fastv, VTW} remove redundant vision tokens to reduce computational overhead.

Finally, in terms of efficient architectures, some work~\cite{moe-llava, mm1} leverages the Mixture of Experts (MoE) architecture to enhance model performance without increasing active parameter counts, while others~\cite{vl-mamba, cobra} adopt efficient Mamba language models as the language backbone.
\section{Additional Experiments}
\label{sec:supp-ae}

        

        

\subsection{W4A16 and W8A8 Results on Large VLMs}
\label{sec:supp-ae-large}
As shown in \cref{tab:main_high}, we present the evaluation results for W4A16 and W8A8 quantized VLMs from the LLaVA-onevision, InternVL2, and Qwen2-VL families.
In most cases, the proposed \methodabbr{} achieves accuracy comparable to the AWQ and SmoothQuant baselines under W4A16 and W8A8 quantization. 
Furthermore, the average accuracy of the quantized VLMs is very close to that of the original FP16 VLMs, indicating that quantization under W4A16 and W8A8 is nearly lossless.

A notable different case arises during the W4A16 quantization of LLaVA-onevision-72B, where AWQ significantly degrades the VLM's accuracy, with the average accuracy falling more than 10\% below that of \methodabbr{} and RTN.
A similar phenomenon also occurs during W3A16 quantization of LLaVA-onevision-72B in \cref{tab:summary-case}, demonstrating that the modality-balancing concept proposed by \methodabbr{} can more consistently maintain high model performance compared to SOTA quantization baselines, whether in high-bitwidth or low-bitwidth quantization scenarios.

\begin{table*}[htbp]
  \centering
  \resizebox{\textwidth}{!}{
      \begin{tabular}{cccccccccc}
        \toprule
        Model & Bitwidth & Method & MMMU & SEED & OCRBench & VizWiz & ScienceQA & TextVQA & Average ($\uparrow$) \\
        \midrule

        ~ & FP16 & - & 46.0 & 74.9 & 62.2 & 60.4 & 85.4 & 76.1 & 67.5\\
        \cmidrule(lr){2-10}
        ~ & ~ & RTN & 44.9 & 74.6 & 61.7 & 59.6 & 89.8 & 75.3 & 67.6\\
        ~ & W4A16 & AWQ & 44.6 & 74.7 & 61.8 & 59.1 & 90.1 & 75.8 & 67.7\\
        LLaVA-onevision-7B & ~ & \methodabbr{} & 44.4 & 74.7 & 62.1 & 59.3 & 90.2 & 75.6 & 67.7\\
        \cmidrule(lr){2-10}
        ~ & ~ & RTN & 46.3 & 74.8 & 63.5 & 60.5 & 90.3 & 75.9 & 68.6\\
        ~ & W8A8 & SQ & 46.0 & 74.9 & 63.2 & 60.7 & 90.3 & 75.7 & 68.5\\
        ~ & ~ & \methodabbr{} & 45.6 & 74.7 & 62.6 & 61.0 & 90.2 & 75.7 & 68.3\\
    
        \midrule

        ~ & FP16 & - & 48.0 & 76.0 & 76.5 & 61.1 & 96.2 & 77.0 & 72.5\\
        \cmidrule(lr){2-10}
        ~ & ~ & RTN & 47.6 & 75.9 & 75.6 & 60.1 & 96.0 & 76.2 & 71.9\\
        ~ & W4A16 & AWQ & 47.1 & 75.8 & 76.7 & 60.1 & 96.3 & 76.4 & 72.1\\
        InternVL2-8B & ~ & \methodabbr{} & 48.9 & 75.9 & 76.7 & 60.8 & 96.3 & 76.5 & 72.5\\
        \cmidrule(lr){2-10}
        ~ & ~ & RTN & 47.4 & 76.2 & 77.3 & 61.0 & 96.2 & 76.9 & 72.5\\
        ~ & W8A8 & SQ & 48.0 & 76.1 & 77.1 & 61.0 & 96.1 & 76.9 & 72.5\\
        ~ & ~ & \methodabbr{} & 48.0 & 76.0 & 77.4 & 61.0 & 96.4 & 77.0 & 72.6\\

        \midrule

        ~ & FP16 & - & 50.6 & 76.4 & 80.7 & 68.3 & 85.1 & 82.0 & 73.8\\
        \cmidrule(lr){2-10}
        ~ & ~ & RTN & 50.2 & 76.0 & 80.1 & 67.4 & 84.5 & 81.2 & 73.2\\
        ~ & W4A16 & AWQ & 50.1 & 76.1 & 80.4 & 68.4 & 85.0 & 81.7 & 73.6\\
        Qwen2-VL-7B & ~ & \methodabbr{} & 50.0 & 76.3 & 80.8 & 68.6 & 84.6 & 81.4 & 73.6\\
        \cmidrule(lr){2-10}
        ~ & ~ & RTN & 49.4 & 76.3 & 80.9 & 68.2 & 84.5 & 81.2 & 73.4\\
        ~ & W8A8 & SQ & 50.1 & 76.3 & 80.6 & 68.5 & 85.0 & 81.5 & 73.7\\
        ~ & ~ & \methodabbr{} & 50.1 & 76.4 & 80.7 & 68.3 & 85.4 & 81.8 & 73.8\\

        \midrule

        ~ & FP16 & - & 56.1 & 78.1 & 73.2 & 69.2 & 90.0 & 79.3 & 74.3\\
        \cmidrule(lr){2-10}
        ~ & ~ & RTN & 56.2 & 77.9 & 72.1 & 68.8 & 90.4 & 78.9 & 74.1\\
        ~ & W4A16 & AWQ & 39.1 & 75.9 & 58.1 & 59.9 & 80.4 & 69.1 & 63.8\\
        LLaVA-onevision-72B & ~ & \methodabbr{} & 56.4 & 77.9 & 72.3 & 69.0 & 90.3 & 79.3 & 74.2\\
        \cmidrule(lr){2-10}
        ~ & ~ & RTN & 56.8 & 78.0 & 73.1 & 69.4 & 90.3 & 79.2 & 74.5\\
        ~ & W8A8 & SQ & 56.3 & 78.0 & 72.7 & 69.2 & 89.7 & 79.0 & 74.2\\
        ~ & ~ & \methodabbr{} & 56.2 & 78.1 & 73.1 & 69.2 & 89.8 & 79.1 & 74.3\\
    
        \midrule

        ~ & FP16 & - & 47.1 & 76.8 & 77.9 & 66.2 & 97.5 & 82.1 & 74.6\\
        \cmidrule(lr){2-10}
        ~ & ~ & RTN & 48.2 & 76.8 & 78.0 & 64.6 & 97.1 & 81.8 & 74.4\\
        ~ & W4A16 & AWQ & 47.4 & 76.8 & 77.1 & 65.9 & 97.3 & 82.0 & 74.4\\
        InternVL2-26B & ~ & \methodabbr{} & 47.2 & 76.8 & 77.5 & 65.4 & 97.5 & 82.1 & 74.4\\
        \cmidrule(lr){2-10}
        ~ & ~ & RTN & 47.4 & 76.5 & 78.4 & 65.1 & 97.3 & 81.7 & 74.4\\
        ~ & W8A8 & SQ & 48.1 & 76.7 & 78.3 & 65.5 & 97.4 & 82.0 & 74.7\\
        ~ & ~ & \methodabbr{} & 47.9 & 76.8 & 78.1 & 66.2 & 97.5 & 82.0 & 74.8\\

        \midrule

        ~ & FP16 & - & 61.1 & 77.6 & 79.9 & 76.0 & 91.6 & 82.5 & 78.1\\
        \cmidrule(lr){2-10}
        ~ & ~ & RTN & 59.8 & 77.7 & 79.6 & 75.8 & 91.3 & 82.6 & 77.8\\
        ~ & W4A16 & AWQ & 61.9 & 77.5 & 79.5 & 75.8 & 92.2 & 82.4 & 78.2\\
        Qwen2-VL-72B & ~ & \methodabbr{} & 60.8 & 77.6 & 79.9 & 75.8 & 92.8 & 82.3 & 78.2\\
        \cmidrule(lr){2-10}
        ~ & ~ & RTN & 60.4 & 77.5 & 79.5 & 75.7 & 90.9 & 82.2 & 77.7\\
        ~ & W8A8 & SQ & 60.3 & 77.5 & 79.7 & 76.1 & 91.3 & 82.0 & 77.8\\
        ~ & ~ & \methodabbr{} & 60.7 & 77.6 & 79.9 & 75.9 & 91.4 & 82.4 & 78.0\\
        
        \bottomrule
      \end{tabular}
    }
  \caption{Additional results on LLaVA-onevision, InternVL2, and Qwen2-VL families. ``SQ'' is short for SmoothQuant.}
  \label{tab:main_high}
\end{table*}

\subsection{Case Studies}
\label{sec:supp-case}

To evaluate the open-ended conversational ability of the quantized VLMs, we evaluate the proposed \methodabbr{} alongside state-of-the-art baselines on the LLaVA-onevision family using conversation benchmarks.
As shown in \cref{tab:summary-case}, we find that \methodabbr{} can significantly outperform SOTA baseline methods when applied to weight-activation quantization for LLaVA-onevision-7B and weight-only quantization for LLaVA-onevision-72B.
Therefore, we analyze the conversation results of VLMs under these two quantization schemes.

Specifically, we manually evaluated the responses of the quantized VLM to each question and identified the following five frequently occurring error types:

\begin{enumerate}
    \item \textbf{No output}: The quantized VLM generates no or only a few valid tokens, as shown in Example~\ref{example:w4a8-2};
    \item \textbf{Randomness}: The quantized VLM randomly generates meaningless symbols, as shown in Example~\ref{example:w3-1};
    \item \textbf{Repetition}: The quantized VLM keeps repeating some certain tokens, as shown in Example~\ref{example:w4a8-1} and Example~\ref{example:w3-4};
    \item \textbf{Condition Missing}: The quantized VLM misses the key points in the questions as shown in Example~\ref{example:w4a8-3};
    \item \textbf{Semantic Error}: The quantized VLM can understand the questions but still generates wrong answers with meaningful and fluent language, as shown in Example~\ref{example:w3-5}.
\end{enumerate}
\vspace{10pt}

We summarize the number of samples corresponding to the above five error types for each quantized VLM, the results are shown in \cref{tab:summary-case}.

For the LLaVA-onevision-7B with W4A8 quantization, \methodabbr{} results in only 45 total bad cases, which is 104 fewer than the 149 bad cases observed with SmoothQuant.
Specifically, the most common type of bad case of SmoothQuant is ``No Output". 
For instance, in Example~\ref{example:w4a8-2}, the W4A8 VLM using SmoothQuant fails to generate any meaningful tokens in response to the provided question. 
In contrast, with \methodabbr{}, the W4A8 VLM experiences only 3 ``No Output" bad cases.
Additionally, while \methodabbr{} leads to more``Semantic Error" cases than SmoothQuant, this is because SmoothQuant often produces outputs that are not even fluently generated, whereas \methodabbr{} enables the quantized VLM to generate fluent responses, thereby increasing the potential for ``Semantic Error" cases.

For the LLaVA-onevision-72B with W3A16 quantization, a similar trend is observed as with the LLaVA-onevision-7B using W4A8 quantization.
\methodabbr{} results in only 38 bad cases, which is 48 fewer than the 86 bad cases reported for AWQ.
Specifically, the most frequent type of bad case for the AWQ baseline is still ``No Output''.
This indicates that AWQ can still lead to VLM failing to generate meaningful and fluent responses in many situations, while MBQ significantly alleviates this type of bad case, narrowing the issues down to ``Semantic Errors''.

\twocolumn[
\begin{convexample}[LLaVA-onevision-7B-W4A8]{label=example:w4a8-1}
\textbf{Question:}
Who painted this artwork, and what is its significance in the art world? 

\begin{center}
\includegraphics[width=\textwidth,height=7cm,keepaspectratio]{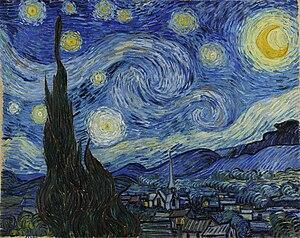} 
\end{center}

\tcbline
\textbf{FP16 Response:}
This artwork was painted by Vincent van Gogh, a post-impressionist painter known for his vibrant use of color and emotional intensity. ``The Starry Night'' is significant in the art world as it exemplifies Van Gogh's unique style and his ability to evoke emotion through his paintings. It has become one of the most recognized and celebrated works of post-impressionism, symbolizing hope and the beauty of the night sky. 

\tcbline
\textbf{SmoothQuant Response:}
This artwork was painted by Vincent van Gogh titled Nightsky Landscape With Bright Starry Sky Above Sleeping Townscape Surrounding Wind Blowing Trees And Hills In Background Van Gogh Postcard Print Poster Blanket Pillowcase Pillow Box Pillow Blanket Pillow Box Pillow Box Pillow Blanket Pillow Box Pillow Box Pillow Box Pillow Box Pillow Box Pillow Box Pillow Box Pillow ow Box Pillow Box Pillow Box Pillow Box Pillow Box Pillow Box Pillow Box Pillow Box Pillow Box Pillow Box Pillow Box Pillow Box Pillow Box Pillow Box Pillow Box Pillow Box Pillow Box Pillow Box Pillow Box Pillow Box Pillow Box Pillow Box Pillow Box Pillow Box Pillow Box Pillow Box Pillow Box Pillow Box Pillow Box Pillow Box Pillow Box Pillow Box Pillow Box Pillow Box Pillow Box Pillow Box Pillow Box Pillow Box Pillow

\textcolor{red}{** Error Type: Repetition.} 

\tcbline
\textbf{\methodabbr{} Response:}
This artwork is painted by Vincent van Gogh, titled ``The Starry Night.'' It's significant in the art world because it exemplifies Van Gogh's unique post-impressionist style and emotional intensity. The swirling patterns in the sky are iconic and have become synonymous with Van Gogh's expressive technique. This painting also signifies his fascination with the cosmos and has been interpreted as an expression of hope amidst darkness. 

\end{convexample}
]

\twocolumn[
\begin{convexample}[LLaVA-onevision-7B-W4A8]{label=example:w4a8-2}
\textbf{Question:}
Describe this photo in detail.

\begin{center}
\includegraphics[width=\textwidth,height=7cm,keepaspectratio]{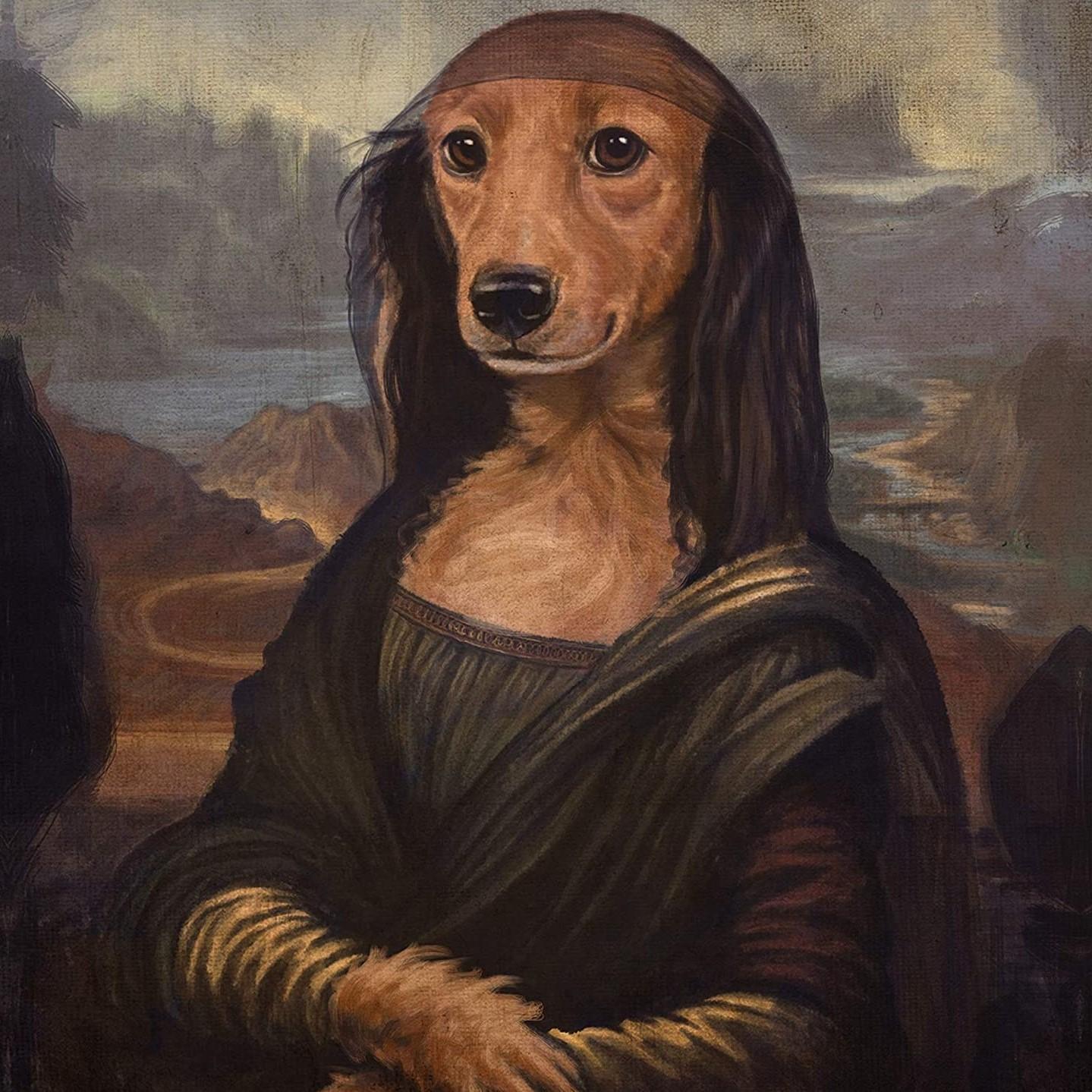} 
\end{center}

\tcbline
\textbf{FP16 Response:}
The image is a digitally created artwork that combines elements of Renaissance art with modern humor. It features a dog's head, specifically that of a dachshund, seamlessly integrated into the iconic pose and attire of Leonardo da Vinci's ``Mona Lisa.'' The dog's face is superimposed onto the original painting, maintaining the Mona Lisa's enigmatic expression and posture. The background remains consistent with the original, depicting a distant landscape with a river and mountains under a cloudy sky. The dog's fur color matches the original's flesh tones, and its ears are styled to resemble the Mona Lisa's hair. The overall effect is a playful fusion of classical art and contemporary animal imagery. 

\tcbline
\textbf{SmoothQuant Response:}
The image presents itself uniquely

\textcolor{red}{** Error Type: No Output. The quantized VLM can only generate a few meaningless tokens.} 

\tcbline
\textbf{\methodabbr{} Response:}
This image is a playful rendition of Leonardo da Vinci's famous painting, the Mona Lisa. However, instead of the original human subject, this version features a dog, specifically a brown dachshund, in place of the Mona Lisa. The dog is portrayed with a slight smile, adding a touch of whimsy to the piece. The background maintains the original's serene landscape of mountains and a calm river. The art style is a fusion of Renaissance painting techniques with a modern twist, reflecting the artist's creative flair. 

\end{convexample}
]

\twocolumn[
\begin{convexample}[LLaVA-onevision-7B-W4A8]{label=example:w4a8-3}
\textbf{Question:}
What is the atmosphere and setting depicted in the image, and how do the elements contribute to creating a relaxing environment? 

\begin{center}
\includegraphics[width=\textwidth,height=7cm,keepaspectratio]{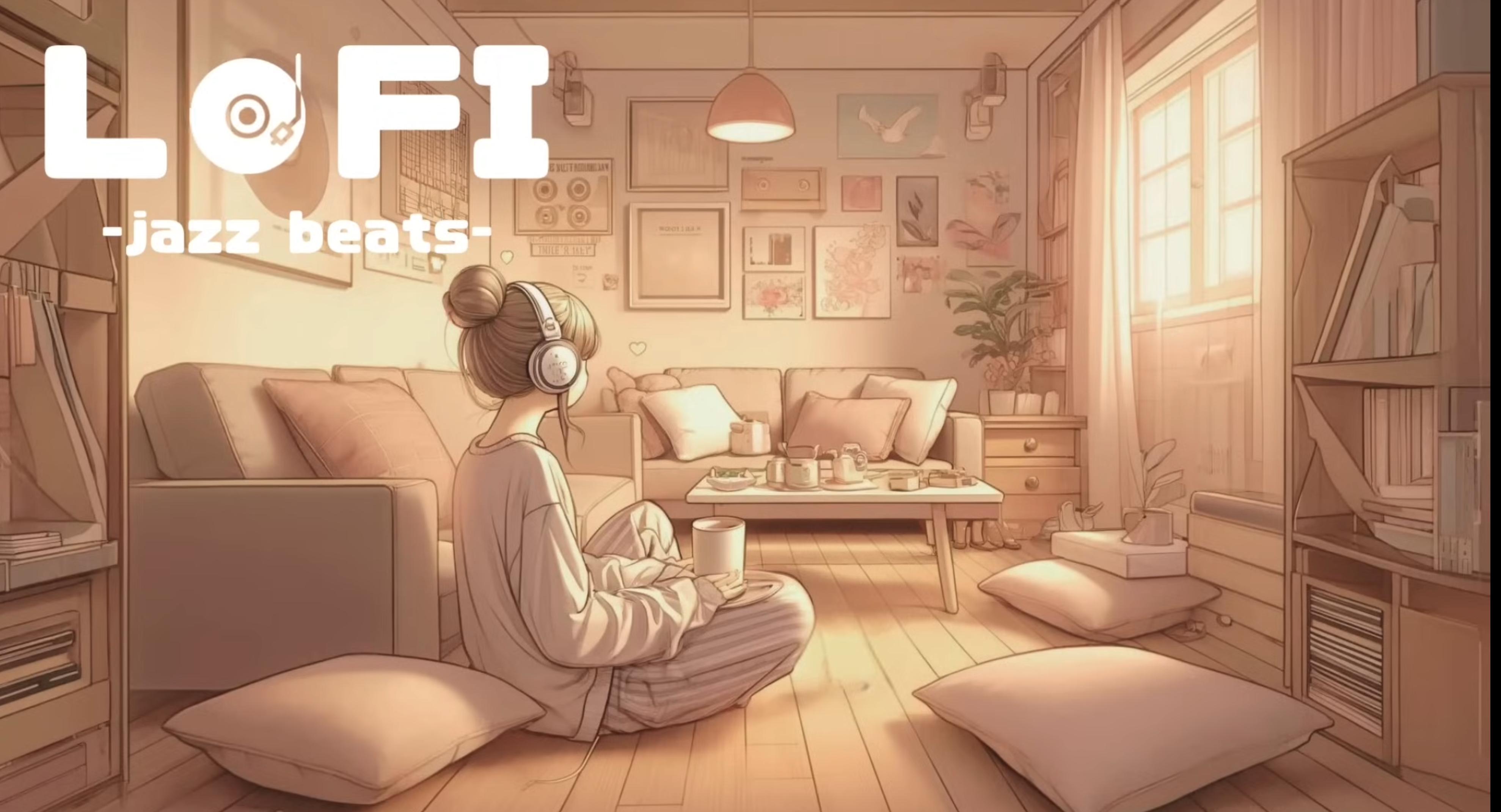} 
\end{center}

\tcbline
\textbf{FP16 Response:}
The atmosphere depicted in the image is one of tranquility and relaxation. The setting is a cozy living room bathed in soft, warm light that suggests it's either dawn or dusk. The presence of the person sitting cross-legged on the floor with headphones on, along with the calm posture and the cup of coffee, implies they are enjoying some quiet time, possibly listening to music. The room's decor, including framed pictures and plants, adds to the homely and peaceful ambiance. The sunlight streaming through the window enhances the serene mood, making it an ideal space for unwinding. 

\tcbline
\textbf{SmoothQuant Response:}
The atmosphere depicted suggests tranquility amidst warmth—an ideal setting conducive to relaxation—one might even say therapeutic listening experience.

\textcolor{red}{** Error Type: Condition Missing. The quantized VLM fails to address the question: ``How do the elements contribute to creating a relaxing environment?''} 

\tcbline
\textbf{\methodabbr{} Response:}
The atmosphere depicted in the image is one of tranquility and relaxation. The setting appears to be a cozy living room designed for comfort and leisure, with warm lighting that suggests either sunrise or sunset. Elements such as plush cushions on the floor, soft lighting from the pendant lamp, and the presence of calming artwork contribute to this relaxing ambiance. Additionally, the person's posture while listening to music through headphones indicates immersion in the music, suggesting that they are enjoying a peaceful moment alone. 

\end{convexample}
]



\twocolumn[
\begin{convexample}[LLaVA-onevision-72B-W3]{label=example:w3-1}
\textbf{Question:}
What model is this car, and can you provide some specifications and performance details about it? 

\begin{center}
\includegraphics[width=\textwidth,height=7cm,keepaspectratio]{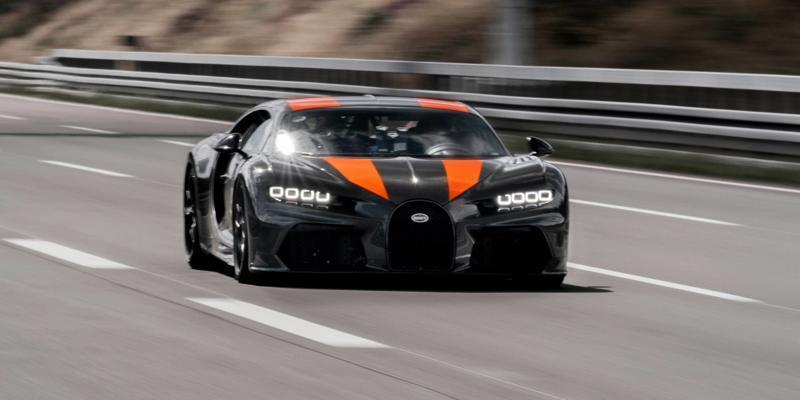} 
\end{center}

\tcbline
\textbf{FP16 Response:}
This car is a Bugatti Chiron, known for its exceptional performance and luxury. It features an 8.0-liter, 16-cylinder engine that produces 1,500 horsepower and 1,300 lb-ft of torque. The Chiron can accelerate from 0 to 60 mph in just 2.4 seconds and has a top speed electronically limited to 261 mph. Its advanced engineering and craftsmanship make it one of the fastest and most exclusive cars in the world. 

\tcbline
\textbf{AWQ Response:}
This is a Bugatti Veyron,EBu,Chirona/EB110,EB110/EB110/EB110/EB110/EB110/EB110/EB110
/EB110/EB110/EB110/EB110/EB110/EB110/EB110/EB110/EB110/EB110/EB110/EB110/EB110/EB110/EB110
/EB110/EB110/EB110/EB110/EB110/EB110/EB110/EB110/EB110/EB110/EB110/EB110/EB110/EB110/EB110
/EB110/EB110/EB110/EB110/EB110/EB110/EB110/EB110/EB110/EB110/EB110/EB110/EB110/EB110/EB110
/EB110/EB110/EB110/EB110/EB110/EB110/EB110/EB110/EB110/EB110/EB110/EB110/EB110/EB110/EB110
/EB110/EB110/EB110/EB110/EB110/EB110/EB110/EB110/EB110/EB110/EB110/EB110/EB110/EB110/EB110
/EB110/EB110/EB110/EB110/EB110/EB110/EB110/EB110/EB110/EB110/EB110/EB110/EB110/EB110/EB110
/EB110/EB110

\textcolor{red}{** Error Type: Randomness. The quantized VLM generates random symbols such as ``EBu'' and ``/EB100''.} 

\tcbline
\textbf{\methodabbr{} Response:}
This is a Bugatti Chiron, known for its incredible performance. It boasts an 8.0-liter, 16-cylinder engine that produces 1,479 horsepower and 1,300 lb-ft of torque. The Chiron can accelerate from 0 to 60 mph in just 2.4 seconds and has a top speed electronically limited to 261 mph. 

\end{convexample}
]


\twocolumn[
\begin{convexample}[LLaVA-onevision-72B-W3]{label=example:w3-5}
\textbf{Question:}
Imagine the fragrance of the fruits in the image. How would you describe this to someone who has never had this fruit before? 

\begin{center}
\includegraphics[width=\textwidth,height=7cm,keepaspectratio]{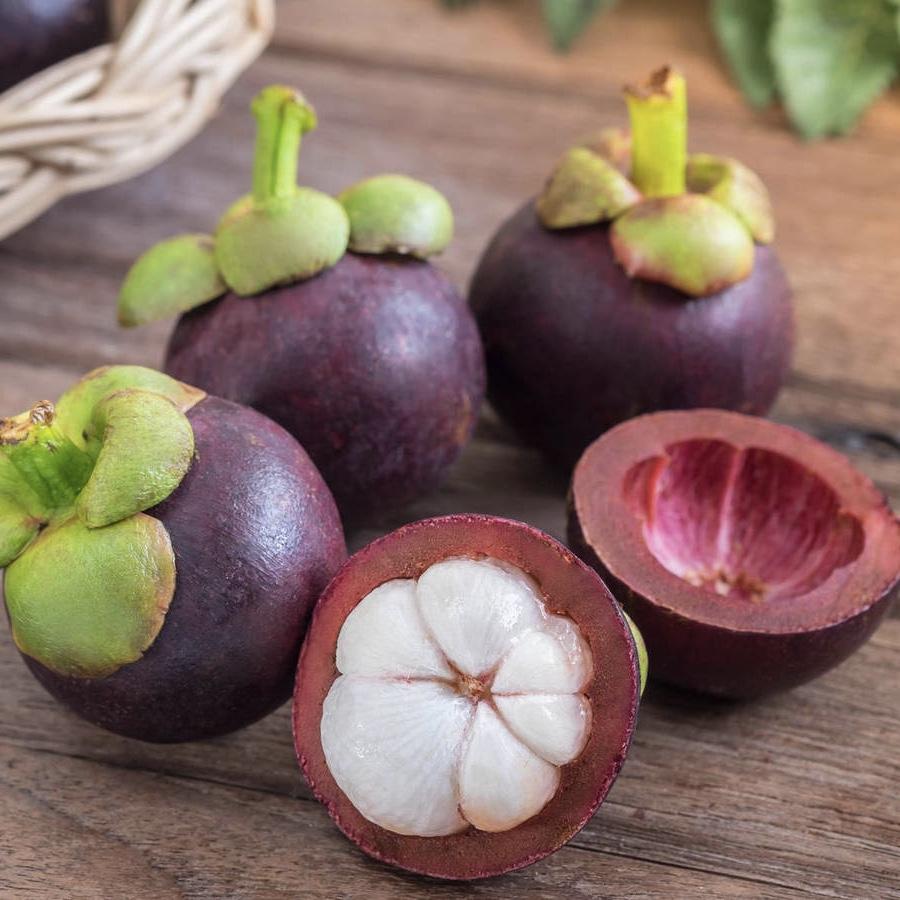} 
\end{center}

\tcbline
\textbf{FP16 Response:}
The fragrance of the mangosteen is a delightful blend of sweet and slightly tart notes, reminiscent of a combination of strawberries, kiwi, and pineapple. It's a fresh and inviting aroma that hints at the juicy and flavorful experience one is about to encounter when tasting the fruit. 

\tcbline
\textbf{AWQ Response:}
The \textcolor{red}{mangos} fruit in the image has a unique fragrance that is both sweet and slightly tart. The sweetness is reminiscent to a combination of honey and a hint of citrus, while the tartness comes from a subtle tang of earthy undertones. When you bite into it, the initial sweetness gives way to a tangy, lingering taste that lingers on your tongue, leaving you with a pleasant and refreshing aftertaste.

\textcolor{red}{** Error Type: Semantic Error. The quantized VLM mistakenly identified mangosteens as mangoes.} 

\tcbline
\textbf{\methodabbr{} Response:}
The fragrance of the mangosteen is a delightful blend of tropical sweetness and subtle floral notes. Imagine a scent that is both refreshing and rich, akin to a combination of strawberries and kiwi with a hint of rose petals. It's a fragrance that promises a burst of flavor and freshness, inviting you to take a bite and experience its unique taste. 

\end{convexample}
]

\twocolumn[
\begin{convexample}[LLaVA-onevision-72B-W3]{label=example:w3-4}
\textbf{Question:}
Write brief HTML/JS to turn this mock-up into a colorful and interactive website, where the joke is replaced by a real joke. 

\begin{center}
\includegraphics[width=\textwidth,height=4cm,keepaspectratio]{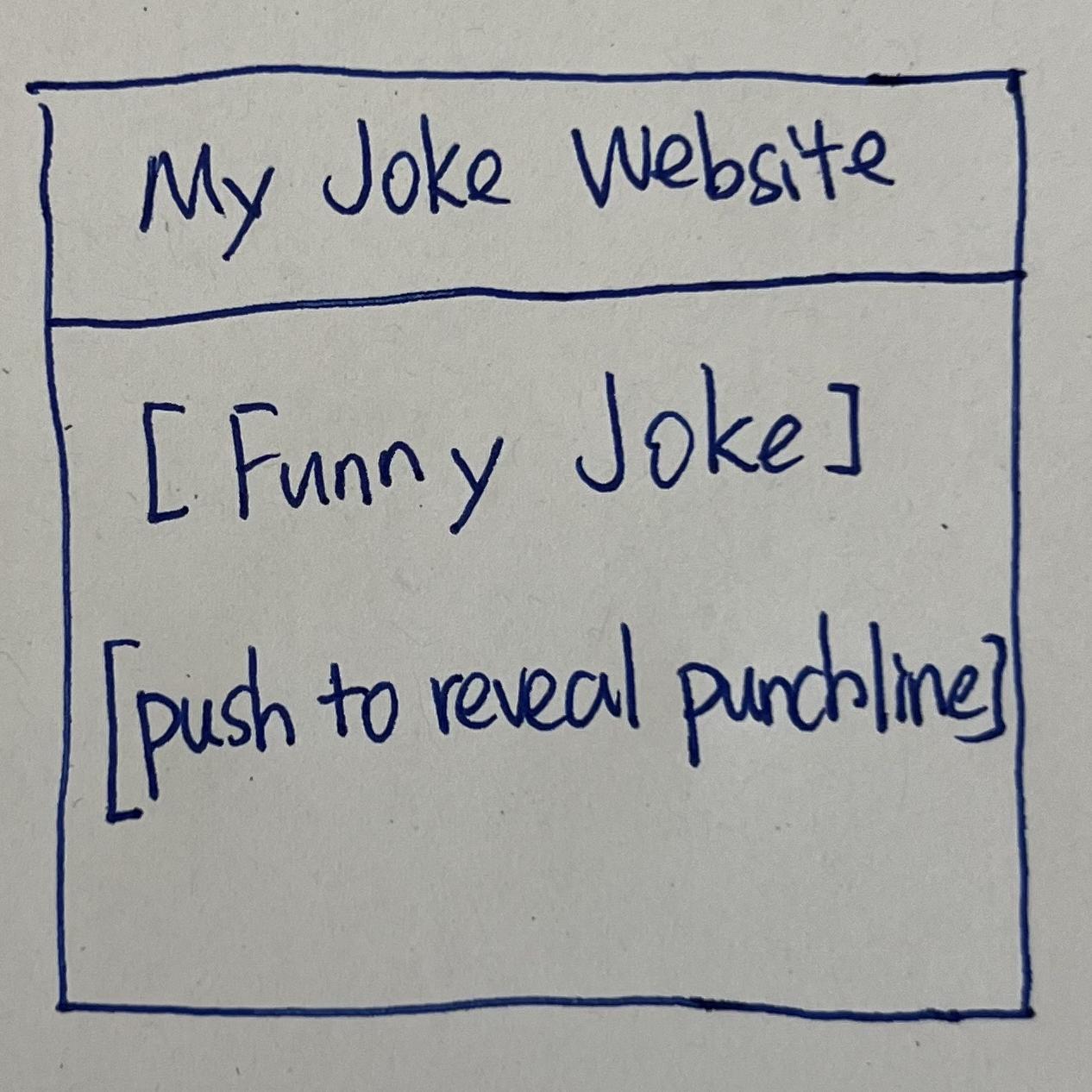} 
\end{center}

\tcbline
\textbf{FP16 Response:}
\begin{center}
\includegraphics[width=\textwidth,height=4cm,keepaspectratio]{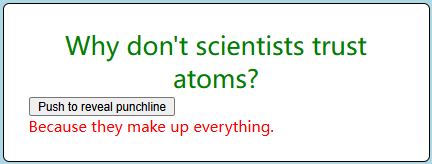} 
\end{center}

\tcbline
\textbf{AWQ Response:}
AS-AS-AS-AS-AS-AS-AS-AS-AS-AS-AS-AS-AS-AS-AS-AS-AS-AS-AS-AS-AS-AS-AS-AS-AS-AS-AS-AS-AS-AS-AS-AS-AS-AS-AS-AS-AS-AS-AS-AS-AS-AS-AS-...

\textcolor{red}{** Error Type: Repetition.} 

\tcbline
\textbf{\methodabbr{} Response:}
\begin{center}
\includegraphics[width=\textwidth,height=4cm,keepaspectratio]{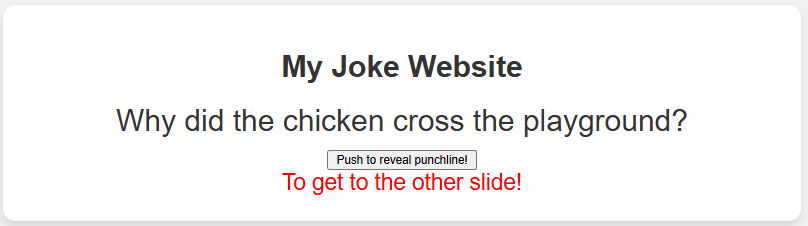} 
\end{center}

\end{convexample}
]


    
    

    
    

\end{CJK}
\end{document}